\documentclass[preprint,1p]{elsarticle}

\usepackage{amssymb}
\usepackage{amsmath}

\usepackage{subcaption}
\usepackage{threeparttable}
\usepackage{booktabs}

\usepackage{fancyhdr}
\journal{Knowledge-Based Systems}

\begin{document}

\begin{frontmatter}



\title{{Brain-Gen: Towards Interpreting Neural Signals for Stimulus Reconstruction Using Transformers and Latent Diffusion Models}
}

\author[label1]{Hasib Aslam\corref{cor1}}
\ead{haslam.bscs21seecs@seecs.edu.pk}

\author[label1]{Muhammad Talal Faiz\corref{cor1}}
\ead{mfaiz.bscs21seecs@seecs.edu.pk}

\author[label1]{Muhammad Imran Malik\corref{cor2}}
\ead{malik.imran@seecs.edu.pk}

\cortext[cor1]{Equal Contribution.}
\cortext[cor2]{Corresponding author.}

\affiliation[label1]{%
    organization={School of Electrical Engineering and Computer Science, National University of Sciences and Technology (NUST)},%
    addressline={H-12},%
    city={Islamabad},%
    postcode={44000},%
    state={Islamabad},%
    country={Pakistan}%
}
\begin{abstract}
Advances in neuroscience and artificial intelligence have enabled preliminary decoding of brain activity. However, despite the progress, the interpretability of neural representations remains limited. A significant challenge arises from the intrinsic properties of electroencephalography (EEG) signals, including high noise levels, spatial diffusion, and pronounced temporal variability. To interpret the neural mechanism underlying thoughts, we propose a transformers-based framework to extract spatial-temporal representations associated with observed visual stimuli from EEG recordings. These features are subsequently incorporated into the attention mechanisms of Latent Diffusion Models (LDMs) to facilitate the reconstruction of visual stimuli from brain activity. The quantitative evaluations on publicly available benchmark datasets demonstrate that the proposed method excels at modeling the semantic structures from EEG signals; achieving up to $6.5\%$ increase in latent space clustering accuracy and $11.8\%$ increase in zero shot generalization across unseen classes while having comparable Inception Score and Fréchet Inception Distance with existing baselines. Our work marks a significant step towards generalizable semantic interpretation of the EEG signals.
\end{abstract}




\end{frontmatter}



\begin{figure}[h]
    \centering
    \includegraphics[width=\textwidth]{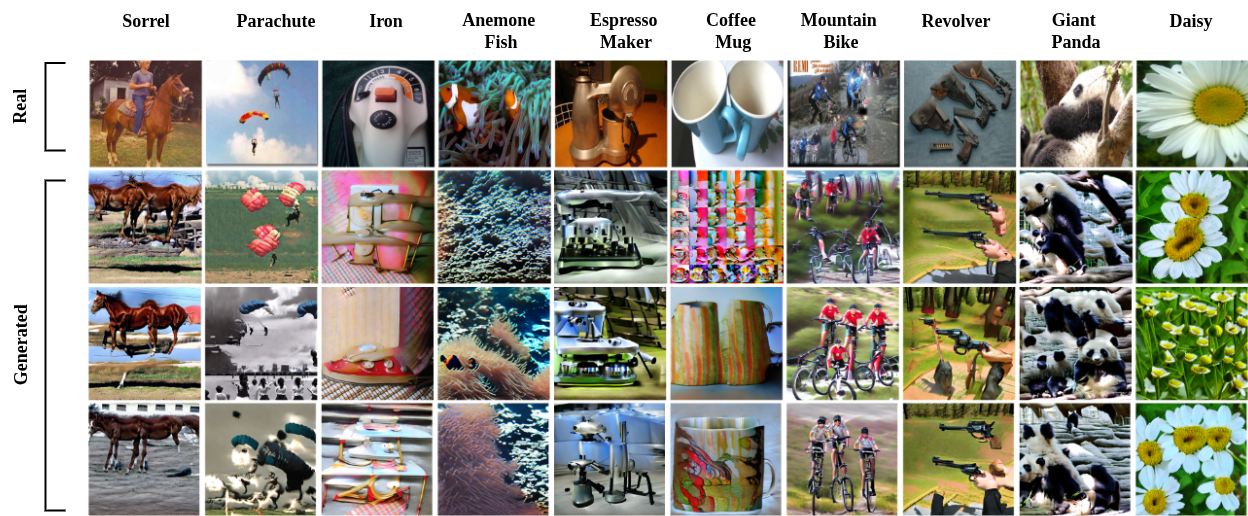}
    \caption{Image generation results from EEG signals corresponding to first 10 classes of EEG-CVPR40 dataset~\ref{ds:eeg-cvpr40} from test split. The first row illustrates ground-truth images shown to the subject with the latter rows consisting of reconstructions, using subject's EEG.}
    \label{img:generation-results}
\end{figure}
\section{Introduction}
The inner working of the human brain is an intriguing open research question, drawing attention of scientists across disciplines, from neuroscience to computer science. Each field approaches this conundrum with their own motivation ranging from advancing biologically plausible computational cognition models to gaining inspiration for development of intelligent machines.

Over the past decade, development in Neuroimaging techniques such as functional Magnetic Resonance Imaging (fMRI) and ElectroEncephaloGraph (EEG) have given us valuable insights into cognitive processes consequently fueling numerous studies~\cite{uugurbil2012development,Nunez2016}.

Computational approaches such as Deep Neural Networks (DNNs) have significantly accelerated these studies. They have proven to be a powerful tool for decoding brain signals and engendering numerous applications in brain-computer interfaces (BCI)~\cite{Schirrmeister2017}. In particular the use of EEG signals has been extended to many medical applications such as assistive technologies~\cite{li2018towards,farwell1988talking}, rehabilitation~\cite{varbu2022past,zhang2018review}, motor function support~\cite{neuper2006motor}, as well as to non-medical domains including drowsiness detection~\cite{Lin2010DrowsinessBCI}, entertainment and gaming~\cite{thomasson2023introduction}.

Moreover, recent studies suggest that brain signals contain cues corresponding to visual and linguistic cognition ~\cite{WEISS2003325,Alday21042019,Gillis10316}. A notable breakthrough demonstrated that brain-signals contain explicit visual cues from thoughts that can be used for reconstructing semantically relevant visuals~\cite{KyotoUniversity2017}. This opens up the possibility of generating images depicting one's thoughts using EEG signals. These findings inspire the objectives of our work.

This research aims to expand the capabilities of visual stimulus reconstruction using EEG data acquired from the cerebral cortex in response to specific image-based visual stimuli. Although most work in this area utilizes fMRI for it's high spatial resolution, EEG was selected for this particular study, as it is more cost-effective and portable method, offering a higher temporal resolution thus capturing the instantaneous bio-electric activity via scalp electrodes. However, EEG presents unique challenges~\cite{Nunez2016}. EEG signals are spatially diffused  resulting in distorted recordings through the skull and scalp. Consequently, accurately locating the signal source is challenging. Lastly, EEG data is inherently noisy, susceptible to movement artifacts, and exhibits a low signal-to-noise ratio~\cite{sadiya2021artifact}, further complicating the decoding process. 

This work leverages the modeling capabilities of deep neural networks to learn semantically rich representations of EEG signals for reconstructing associated visual stimuli. To address the apparent disparity between the two data modalities involved - static, two-dimensional visual stimuli and dynamic, multi-channel EEG time-series signals, we employ a two-stage framework. The key steps are: (i) train a deep neural network to learn EEG signal representations and (ii) reconstruct the image using a diffusion model guided by the learned EEG latents. In this paper, we make the following contributions:

\begin{itemize}
    \item We propose a largely unexplored transformer based architecture with explicit focus on both spatial and temporal dynamics of EEG signals, which enhances the semantic structure of our latent space.
        
    \item We test the proposed framework on benchmark datasets for downstream tasks of classification, clustering, stimulus reconstruction and class generalization, where it outperforms most existing baselines.

\end{itemize}

\section{Related Work}

This section reviews the recent approaches relevant to generating images from EEG signals. Analogous to the standard structure for text-to-image generation~\cite{hussen2025advancing}, our literature review and consequently the EEG-to-Image reconstruction pipeline consists of two core processes: (i) Encoding raw signals into semantic representations and (ii) Decoding these embeddings into realistic images.
This two-fold review serves as the backdrop for understanding the structure and rationale of our approach.

\subsection{Encoding EEG Signals}
Depending on the application domain and the scope of analysis,
EEG signal interpretation techniques can generally be categorized into
two broad approaches. The first approach frames EEG interpretation as a classification problem, wherein the objective is to assign
each signal to a corresponding class label. Representative studies include~\cite{spampinato2017deep,zheng2020decoding,khare2022neurovision,jiang2020brain}.
Secondly, it is also common to explicitly extract features from EEG signals at this stage and use them for down stream applications~\cite{tirupattur2018thoughtviz,singh2023eeg2image,singh2024learning, bai2023dreamdiffusion}. These aforementioned works have either used contrastive learning~\cite{tirupattur2018thoughtviz,singh2023eeg2image,singh2024learning}, or masked signal modeling~\cite{bai2023dreamdiffusion} as training objectives. While it is also possible to use a combination of both methods~\cite{mishra2021eeg}.
Regardless of the problem setup, most of the stated studies have either used Convolutional Neural Networks~\cite{krizhevsky2017imagenet,mishra2021eeg} or LSTMs, as in ~\cite{spampinato2017deep, jiang2020brain, singh2023eeg2image, khare2022neurovision}, or combination of both~\cite{zheng2020decoding}, as backbone architectures of their encoders, while some such as~\cite{bai2023dreamdiffusion} have used the architecture similar to ViTLarge~\cite{dosovitskiy2020image} as the encoder backbone.

\subsection{Decoding Image 
        From Encoded EEG}
Reconstruction of visual stimuli from EEG signals is commonly achieved through generative models such as Generative Adversarial Networks (GANs) and Variational Autoencoders (VAEs). These models learn a conditional map from the latent space of EEG features to the pixel space of corresponding images~\cite{kavasidis2017brain2image}. However, the scarcity of paired EEG-image data limits this approach as effectively training generative models requires a lot of data. 

To enhance image generation quality, GAN variants such as cProGAN~\cite{karras2018progressivegrowinggansimproved} with additional convolution layers have been used, improving image resolution~\cite{khare2022neurovision}. DCLS-GAN is another variant that produces images from both noise vector as well as the combined description vector~\cite{fares2020brain}. EEGStyleGAN-ADA proposed by Singh et al.~\cite{singh2024learning} extends~\cite{karras2020training} by incorporating adaptive augmentation on images before passing them to discriminator to overcome the deficiency of large datasets. In addition to the widespread use of GANs, several works \cite{bai2023dreamdiffusion,zeng2023dm,qian2024neurodm,lopez2025guess} have employed diffusion models \cite{ho2020denoising} for image generation, achieving superior quantitative and qualitative performance.

\begin{figure*}[h]
    \centering
    \includegraphics[width=\textwidth]{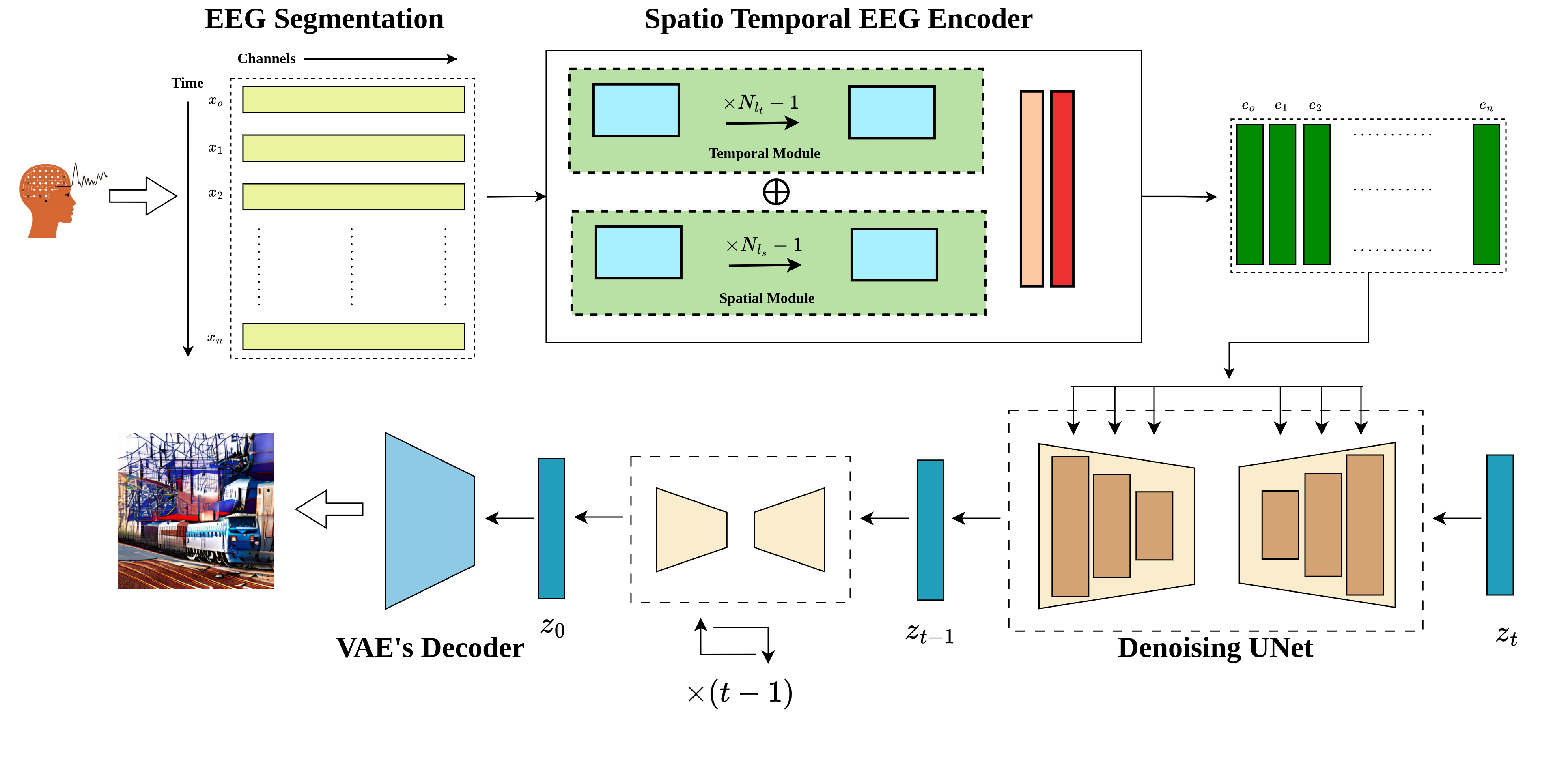}
    \caption{System Diagram: First we use sliding window segmentation to extract multiple, sub samples from an EEG signal. Each of the sub sample is then encoded independently by the spatio-temporal encoder. The resulting sequence of EEG representations is then used to condition the reverse diffusion process of the Stable Diffusion, resulting in semantically aligned reconstruction of the visual stimuli. 
    }
    \label{img:system-overview}
\end{figure*}

\section{Methodology}

This section presents our proposed framework for learning semantically rich representations of EEG signals and generating relevant images. We use a conditioning mechanism that integrates EEG latents into the diffusion process, generating images and forming our reconstruction pipeline. Below, we detail the problem statement, key components and individual processes for this task.

\subsection{Problem Statement}
Let set \( D \) be a collection of \( n \) unique pairs of EEG signals, images, and corresponding images class labels, represented as
\begin{displaymath}
    D = \{(x_i, q_i ,y_i)\}_{i=1}^{n}
\end{displaymath}
such that \( x_i \in \mathbb{R}^{t \times c} \) , \( y_i \in \mathbb{R}^{w \times h \times 3} \) and \( q_i \in \mathbb{R} \)
where \( c, t \) represents the number of channels and number of time steps in the EEG signal, respectively and \( w, h \) represent the width and height of the original image.  
Our goal is to find a parameterized function \( f_\theta \) that maps \( x_i \) to \( y_i \) for \(  i = 1, \dots, n \), where $\theta$ denotes the learnable parameters. 

Simply, our objective is to use EEG signals recorded while exposing subjects to visual stimuli and achieve semantically aligned regeneration of original images.

\begin{figure}[ht]
    \centering

    \begin{subfigure}[t]{0.49\linewidth}
        \centering
        \includegraphics[height=5cm]{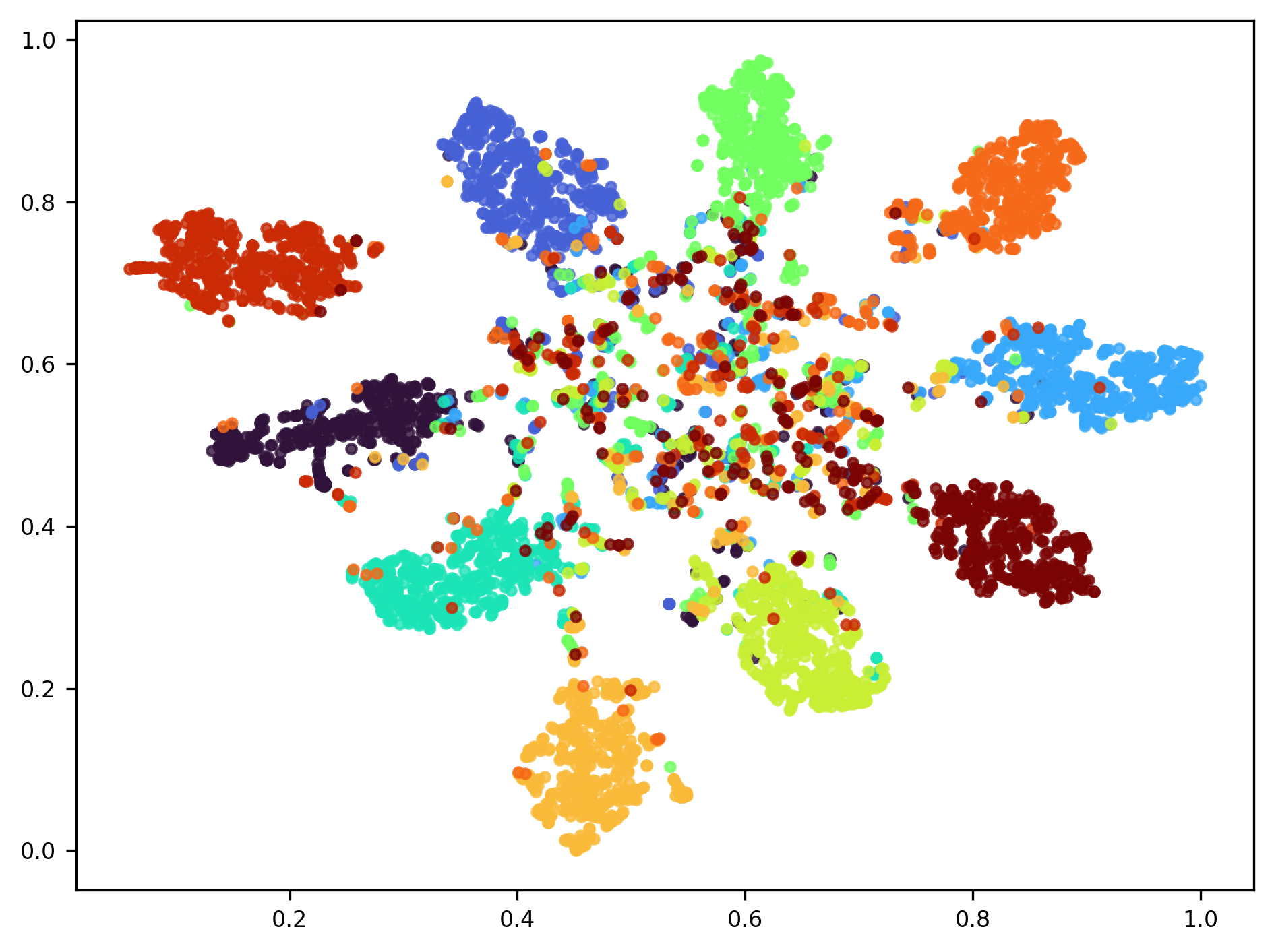}
        \caption{Thought Viz}
    \end{subfigure}
    \hfill
    \begin{subfigure}[t]{0.49\linewidth}
        \centering
        \includegraphics[height=5cm]{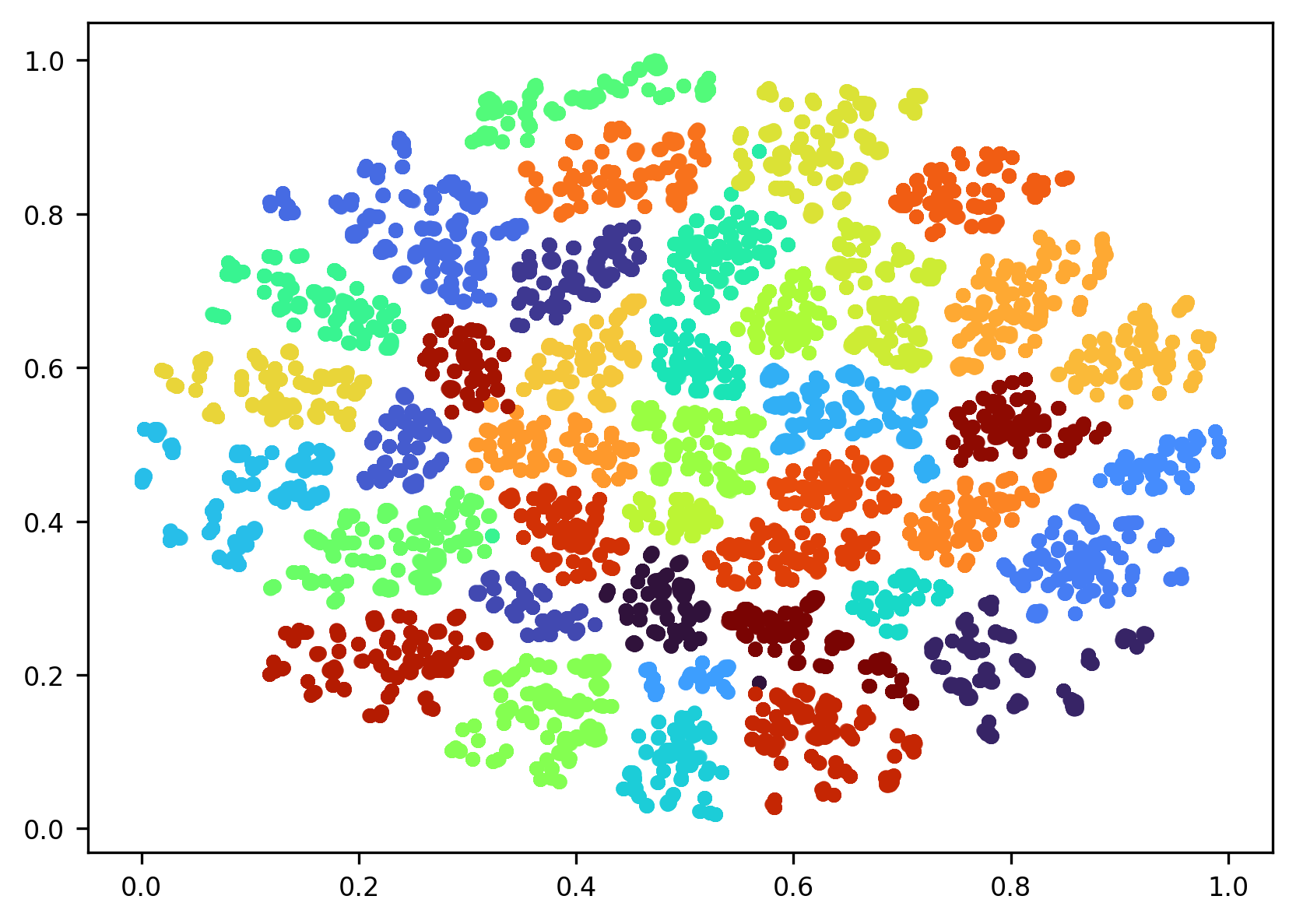}
        \caption{EEG-CVPR 40}
        \label{fig:second}
    \end{subfigure}

     \caption{Visualization of features extracted by proposed encoder from unseen samples from ThoughtViz dataset in left column and EEG-CVPR40 in right column using TSNE plots.}
    \label{fig:tsne}
\end{figure}

\subsection{Spatio-Temporal EEG Encoder}
\label{sec:Saption-Temporal-Encoder}

The first component of our proposed framework is a Spatio-Temporal encoder. To obtain feature rich embeddings we create a design that (i) captures long-range temporal dependencies, (ii) models spatial relationships across channels, and (iii) preserves both local and global contextual information. Existing approaches rely predominantly on CNNs and RNN variants such LSTMs, each with inherent limitations for this objective.

CNNs are primarily designed to extract hierarchal features from grid structures such as images. Their application on time-series data such as EEG requires rearranging the signal into a 2-dimensional grid, with channels on one axis and discrete time samples on the other. Next, successive filters of varying sizes are applied on the grid to extract spatial information. As the filters operate on localized patches in this grid, their ability to capture dependencies across time or channels becomes limited. Consequently, capturing global dependencies requires large receptive fields and deep filter stacks whose computational cost scales non-linearly with model depth and input length.

In contrast, RNNs in general and specifically LSTMs are well suited for processing time signals because of their ability to capture long term dependencies. At each time step RNN units can access discrete values from all channels, when the channel dimension is considered as the feature dimension thus learning the necessary dependencies. However, the sequential processing in RNNs lacks an explicit mechanism for attending to all time steps simultaneously, leading to limited global temporal awareness.

To solve the issues discussed above, we take our inspiration from work on EEG based authentication systems~\cite{du2022eeg}, where transformer based architecture, incorporating explicit spatial and temporal modules, is employed for classification of the subject's EEG signal. 
We adapt the proposed architecture to a contrastive learning framework, learning discriminative features instead of class labels. 
Through experiments we establish the sufficiency of the temporal module for low complexity, downstream tasks such as that of classification and clustering.
However, for stimulus reconstruction, we observe significant performance gains when the spatial module is also included. \textit{See~\ref{results} for more}. The general architecture of the encoder is given below.

\subsubsection{Temporal Encoder}
The temporal encoder processes EEG signals across time, preserving temporal dependencies. The temporal encoding of an input $x_i \in \mathbb{R}^{t \times c}$ is computed as:

\[
x_t = {E_t}(x_i)
\]
\[
E_t: \mathbb{R}^{t \times c} \to \mathbb{R}^{t \times c}
\]

where $E_t(\cdot)$ represents a multi-layered, multi-head self-attention transformer block, with $N_{l_t}$ layers, each with $N_{h_t}$ heads, $c$ is the number of electrodes used for recording EEG and $t$ defines the number of time steps in the sample.

\subsubsection{Spatial Encoder}
The spatial encoder captures correlations among EEG channels by applying a transformer-based model along the channel dimension. Given the feature encoding from the temporal encoder $x_t \in \mathbb{R}^{t \times c}$ the representation obtained is as follows:

\[
x_s = E_s(x_t^T)
\]
\[
E_s: \mathbb{R}^{c \times t} \to \mathbb{R}^{c \times t}
\]

where $x^{T}$ represents the transpose of $x$ and $E_s(\cdot)$ represents a multi-layered, multi-head self-attention transformer block, with $N_{l_s}$ layers, each with $N_{h_s}$ heads, $t$ is the sequence length, and $c$ is the number channels in EEG signal.

\subsubsection{Latent Projection}
The final representation is obtained by flattening the combined spatial and temporal embeddings and projecting them into a lower-dimensional latent space using a fully connected network.

\[
z = F(x_s)
\]
\[
F: \mathbb{R}^{t \times c} \to \mathbb{R}^{1 \times d }
\]

where $F$ consists of fully connected layers with dropout regularization, preceded by the flattening operation. So in its final form our encoder is defined as $E_{\theta}(\cdot) = F(E_s(E_t(\cdot))) $ such that $E_\theta: \mathbb{R}^{t \times c} \to \mathbb{R}^{1 \times d }$

\section{Contrastive Learning}
\label{sec:contrastive-learning}

We optimize the encoder using a contrastive objective based on triplet loss~\cite{schroff2015facenet}. The goal is to learn representations that remain consistent across EEG signals corresponding to the same image while separating representations from different images. We define the margin-based triplet loss as:
\[
\mathcal{L}_{\text{triplet}} =
\max\!\left(
d(E_\theta(x_a),\, E_\theta(x_p))
-
d(E_\theta(x_a),\, E_\theta(x_n))
+
\beta,\;
0
\right),
\qquad
\beta = 0.05,
\]
where \( q_a = q_p \) and \( q_a \neq q_n \), and \( d(\cdot,\cdot) \) denotes cosine distance.

To improve training stability and promote better separation between classes, we employ semi-hard negative mining, selecting negatives that satisfy
\begin{align*}
d(E_\theta(x_a), E_\theta(x_p))
&< d(E_\theta(x_a), E_\theta(x_n)) \\
&< d(E_\theta(x_a), E_\theta(x_p)) + \alpha,
\end{align*}
with \( \alpha = 0.1 \). Selecting negatives that are harder than easy examples, but still lie within the margin defined by the anchor to positive distance.

\subsection{EEG Signal Tokenization}
\label{sec:tokenization}
To make the feature extracted from EEG signals compatible with the conditioning mechanism of stable diffusion, which expects sequence of embeddings from textual prompt, we use the sliding window approach to extract multiple sub samples, each of which is treated as an individual token and is processed independently. 
Let an EEG recording be $x \in \mathbb{R}^{t \times c}$, where \(t\) is the number of time steps and \(c\) the number of channels.
We define a sliding-window extraction operator with window length \(l\) and stride \(s\) where the number of extracted segments $n$ is:
\[
n = \left\lfloor \frac{t - l}{s} \right\rfloor + 1.
\]
Each windowed segment is
\[
x_i = x[t_i : t_i + l - 1] \in \mathbb{R}^{l \times c},
\qquad i = 1,\dots,n,
\]
with
\[
t_i = (i-1)s.
\]
Each segment is independently mapped by the EEG encoder
\[
E_\theta : \mathbb{R}^{l \times c} \to \mathbb{R}^{d}.
\]
Thus, each token embedding is
\[
e_i = E_\theta(x_i) \in \mathbb{R}^{1 \times d}.
\]
Collecting all embeddings yields the final token sequence
\[
S_t = \big[ e_1, e_2, \dots, e_n \big] \in \mathbb{R}^{n \times d}.
\]

\subsection{Image Generation}
\label{sec:image-geneartion}
We employ Stable Diffusion-2~\cite{Rombach_2022_CVPR} for image reconstruction task, through conditioning the reverse diffusion process on features extracted from EEG signals using the cross-attention mechanism~\cite{vaswani2017attention}. Specifically we use the $S_t \in \mathbb{R}^{n \times d}$, the sequence of EEG features obtained from tokenization based approach described above, to replace the embedding sequence from textual prompt in traditional Stable Diffusion-2~\cite{Rombach_2022_CVPR} pipeline. Here the cross-attention is calculated as:

\[
\operatorname{Attention}(Q,K,V) = \operatorname{softmax}\!\left(\frac{QK^\top}{\sqrt{d}}\right)V,
\]
where the learnable projection matrices 
\[
W^{(i)}_Q \in \mathbb{R}^{d \times h}, \quad W^{(i)}_K \in \mathbb{R}^{d \times h}, \quad \text{and} \quad W^{(i)}_V \in \mathbb{R}^{d \times h}
\]
are used to compute
\[
Q = W^{(i)}_Q \cdot \phi_i(z_t), \quad K = W^{(i)}_K \cdot S_t, \quad V = W^{(i)}_V \cdot S_t.
\]
Here, $\phi_i(z_t) \in \mathbb{R}^{N \times d_i}$ denotes the intermediate representation of the UNet implementing $\phi_\theta$.
Based on EEG-Image conditioning pairs, we then learn the conditional LDM via
\[
\mathcal{L}_{\text{LDM}} := \mathbb{E}_{x,y,\epsilon \sim \mathcal{N}(0,1)}\left[
\left\|\epsilon - \epsilon_\theta\big(z_t, t, S_t\big)\right\|_2^2
\right], \tag{3}
\]
where $S_t$ is obtained from EEG encoder. To prevent overfitting of the EEG encoder for the image generation process. we use it's weights from the contrastive learning task and keep them frozen. We finetune only the denoising UNet $\epsilon_\theta$, so that it learns the semantics in the latent space of EEG encoder. We start the training from the publicly available weights of Stable Diffusion 2.

\section{Experiments and Results}
In this section we discuss our experiments in detail. Alongside quantitative and qualitative results, we also compare with previous works to show the efficacy of our method.
\subsection{Datasets}
\subsubsection{EEG-CVPR40}
\label{ds:eeg-cvpr40}
 This dataset introduced in~\cite{spampinato2017deep} consists of images from $40$ ImageNet~\cite{deng2009imagenet} classes where $50$ samples of images are drawn from each class. The resulting $2000$ samples are shown to each of six different subjects for $0.5$ seconds, and the EEG signals are recorded at a sampling rate of $1$ kHz. It results in $12000$ samples in total, out of which $36$ samples were discarded by the original authors. They also cut down the remaining $11964$ samples to $440$ time steps.
To keep our work comparable, we use the raw dataset with the same train, validation, and test splits as provided in the original work. Finally in our work, we use the sliding window approach to create multiple sub-sample from each original sample. This is done independently in each of train, validation and test splits to ensure that there is no data leakage. 
\subsubsection{Thought Viz}
\label{ds:thoughtviz}
This dataset first used in~\cite{tirupattur2018thoughtviz} for EEG to Image generation task is a limited dataset in which $23$ subjects were shown images belonging to $10$ different classes of ImageNet~\cite{deng2009imagenet} and the EEG signals using $14$ channels were recorded at sampling frequency of $2048$ Hz and later down sampled to $128$ Hz. The publicly available dataset contains $45390$ train and $5706$ test samples each containing $32$ time steps from $14$ channels, paired with class labels of original images that were shown to the subjects.
Unlike EEG-CVPR40, this dataset only contains EEG signal to image class mappings. Consequently, due to the lack of pixel level information about the original image, this dataset is only used to test the generalization abilities of our proposed EEG encoder and not for image generation.
\subsection{Evaluation Metrics}
\subsubsection{EEG K-Means Clustering Accuracy}
\label{sec:metrics-KMean}

To assess the strength of the features extracted by our \textit{Spatio-Temporal Encoder}, we consider the accuracy of k-means clustering over features extracted from EEG signals. We take \textit{k}, i.e, number of means to be the same as the number of classes in the dataset which are $40$ and $10$ in EEG-CVPR40 and ThoughtViz datasets respectively. 

\subsubsection{Zero Shot Class Generalization}
To assess the generalization abilities of the proposed encoder across different classes, we train the encoder on a subset of classes and examine its feature extraction on remaining classes using k-means clustering, where $k$ is set to the number of held out classes. To make our work comparable we keep the experimental setting similar to existing works~\cite{Cogni-Net-8803717,singh2024learning} where for training the encoder, samples from 34 classes are used and during testing k-means clustering is performed on the features extracted from samples belonging to the 6 held out classes.

\subsubsection{Classification Accuracy}
\label{sec:metrics-CLS}
Even though, the proposed approach is based on a contrastive learning framework with the final goal being stimulus reconstruction, we evaluate the proposed encoder on the downstream task of predicting the class labels of the images shown to the subject corresponding to the EEG signal (classification).  
After the proposed encoder has been trained using the contrastive learning objective, a classification head consisting of full connected layers is added on top of the trained encoder. This head is trained using cross-entropy loss while keeping the weights of the base encoder frozen.

\subsubsection{Inception score (IS)} 
\label{sec:metrics-IS}
To estimate the quality of images generated from EEG signals, we use the Inception Score~\cite{salimans2016improved}, which provides a measure of the distinctiveness and sharpness of generated images. Using logits from Inception-V3~\cite{szegedy2016rethinking}. The Inception Score is defined as:
\begin{equation}
\text{IS}(G) = \exp\left( \mathbb{E}_{\mathbf{x} \sim p_g} \left[ D_{\mathrm{KL}}(p(y|\mathbf{x}) \| p(y)) \right] \right),
\end{equation}
where $\mathbf{x}$ is a generated image, $p(y|\mathbf{x})$ is the conditional label distribution predicted by the Inception network, and $p(y) = \int p(y|\mathbf{x}) p_g(\mathbf{x}) d\mathbf{x}$ is the marginal distribution over all generated images. A higher IS indicates that the generated images are both diverse (i.e., $p(y)$ has high entropy) and meaningful (i.e., $p(y|\mathbf{x})$ is peaked).

\subsubsection{Fréchet Inception Distance (FID)}  
\label{sec:metrics-FID}
To measure the similarity between the distribution of generated images and real ones, we use Fréchet Inception Distance (FID)~\cite{NIPS2017_8a1d6947}. We use the pre-trained Inception-V3~\cite{szegedy2016rethinking} model’s final pooling layer to extract a $2048$-dimensional feature vector for each image. The FID is then computed between the multivariate Gaussians fitted to these feature representations of real and generated images as follows:

\begin{equation}
\text{FID}(r, g) = \|\mu_r - \mu_g\|_2^2 + \mathrm{Tr}\left(\Sigma_r + \Sigma_g - 2\left(\Sigma_r \Sigma_g\right)^{1/2} \right),
\end{equation}
where $\mu_r$ and $\Sigma_r$ are the mean and covariance of the features for real images, and $\mu_g$ and $\Sigma_g$ are those for generated images. A lower FID indicates that the generated images are more similar to the real ones in terms of both quality and diversity.
\begin{figure}[htbp]
    \centering
    \includegraphics[height=6cm]{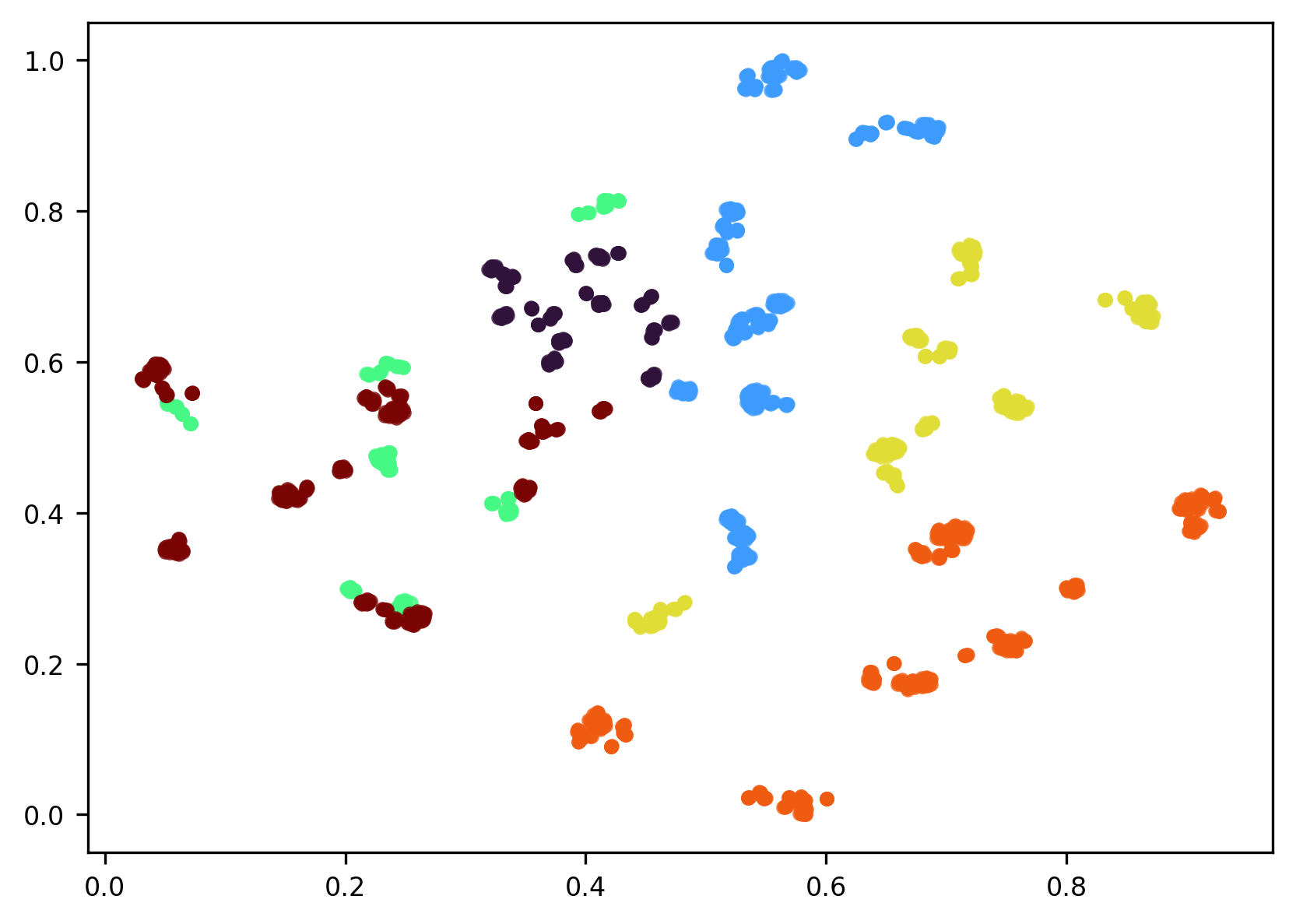}
    \caption{Visualization of features extracted by the proposed encoder from samples of unseen classes from EEG-CVPR40 using t-SNE.}
    \label{fig:tsne-unseen}
\end{figure}
\begin{table}
\centering
\caption{Classification and K Means Accuracies on ThoughtViz and EEG-CVPR40}
\label{tab:acc_comparison}
\resizebox{\columnwidth}{!}{
  \begin{tabular}{lcc}
    \toprule
    \textbf{Model} & \textbf{Classification Acc (\%)} & \textbf{K Means Acc (\%)} \\
    \midrule
    \multicolumn{3}{c}{\textbf{ThoughtViz}} \\
    \midrule
    ThoughtViz \cite{tirupattur2018thoughtviz} & 0.72 & 0.18 \\
    SiameseCNN \cite{mishra2021eeg} & 0.779 & -- \\
    EEG2Image \cite{singh2023eeg2image} & 0.55 & 0.52 \\
    EEGStyleGAN-ADA \cite{singh2024learning} & 0.74 & 0.721 \\
    \textbf{Ours} & \textbf{0.767} & \textbf{0.768} \\
    \midrule
    \multicolumn{3}{c}{\textbf{EEG-CVPR40}} \\
    \midrule
    LSTM Encoder \cite{spampinato2017deep} & 0.829 & 0.45 \\
    DML \cite{jiang2019context} & 0.977 & -- \\
    LSTM-CNN \cite{zheng2020decoding} & 0.944 & -- \\
    BioLSTM \cite{jiang2020brain} & 0.991 & -- \\
    NeuroVision \cite{khare2022neurovision} & 0.988 & -- \\
    EEGStyleGAN-ADA \cite{singh2024learning} & 0.983 & 0.961 \\
    \textbf{Ours} & \textbf{0.994} & \textbf{0.995} \\
    \bottomrule
  \end{tabular}
}
\end{table}

\subsection{Results}
\label{results}

\subsubsection{EEG Encoding Results on ThoughtViz and EEG-CVPR40}
In Table~\ref{tab:acc_comparison}, classification accuracy and k-means accuracy of EEG signals from test splits of both datasets is presented.
Relative to existing baselines our proposed encoder achieves an increase of $6.51 \%$ and $3.53 \%$ in the k-means accuracy on ThoughtViz and EEG-CVPR40 dataset respectively. For EEG-CVPR40 dataset the proposed encoder achieved highest classification accuracy $0.994$, while the classification results on ThoughtViz remain comparable to the best performing architecures..

For EEG-CVPR40 dataset~\ref{ds:eeg-cvpr40}, the results were obtained by training the proposed architecture~\ref{sec:Saption-Temporal-Encoder} with $N_{l_t}=2$ and $H_{l_t}=4$ and $N_{l_s}=0$, \textit{using only the temporal module}, with sequence length of $t=32$ on contrastive learning objective, with hidden dimension $d=128$, for $1024$ epochs, with constant learning rate of $3e-5$ using the Adam optimizer. The choice of hyper-parameters for encoder modules and signal length used are discussed in~\ref{sec:contribution-spatio-temporal} and~\ref{sec:time-analysis} respectively.

For ThoughtViz dataset~\ref{ds:thoughtviz}, the results were obtained by training the proposed architecture~\ref{sec:Saption-Temporal-Encoder} with $N_{l_t}=2$ and $H_{l_t}=2$ and $N_{l_s}=0$, \textit{using only the temporal module}, with sequence length of $t=32$ on contrastive learning objective, with hidden dimension $d=128$, for $4096$ epochs, with constant learning rate of $3e-5$ using the Adam optimizer.
\subsubsection{Zero Shot Class Generalization}
Table~\ref{tab:unseen_cvpr40} shows the performance of proposed encoder on class generalization task on EEG-CVPR40 dataset and its comparison with existing baselines. Our methods achieved k-means clustering accuracy of $69.9\%$ on features extracted from unseen classes, which is $11.84\%$ higher, relative to previously reported results. Moreover for comparison, KNN accuracy is computed where relative gain of $15.11\%$ is observed.

The results were obtained by training the proposed architecture~\ref{sec:Saption-Temporal-Encoder} with $N_{l_t}=2$ and $N_{h_t}=4$ and $N_{l_s}=0$, \textit{using only the temporal module}, with sequence length of $t=32$ on contrastive learning objective, with hidden dimension $d=128$, for $60$ epochs, with constant learning rate of $3e-5$ using the Adam optimizer and testing on remaining 6 classes, the excluded set of classes was kept same as in~\cite{Cogni-Net-8803717,singh2024learning}. 
\subsubsection{Image Generation Results on EEG-CVPR40}
\label{res:img-gen-cvpr40}
The quantitative results of proposed methodology on EEG-CVPR40 dataset are presented in Table~\ref{tab:fid_is_cvpr40} while qualitative results displayed in Figure~\ref{img:generation-results}.
Comparable to existing baseline~\cite{lopez2025guess}, the proposed method achieves Inception Score of $25.15$ and Fréchet Inception Distance of $81.07$.


We stress upon the need of both IS and FID for the accurate assessment of the stimulus reconstruction. IS is important as it is the measure of the visual quality of the generated images. The low values of IS are indication of the lack of diversity in generated population of images or the lack of semantic clearity in them, however any image generation model capable of generating high quality, diverse images will have high IS, regardless of their relevance with original visual stimuli. 
That's why it is necessary to define some metric, like FID that measures semantic similarity between the distribution of generated and real images. Smaller FID means that the generated images are closer to original ones and vice versa.

The results discussed above are obtained by finetuning the Stable Diffusion on EEG-CVPR40 dataset~\ref{ds:eeg-cvpr40} following the method explained in~\ref{sec:image-geneartion}. While finetuning, the generation was conditioned on features extracted through tokenization encoding of the corresponding EEG signals~\ref{sec:tokenization}. We used the constant learning rate of $1e-5$ and finetuned the model for $25000$ steps. Notably, we observe that image generation quality increases with increase in complexity of encoder. \textit{See Appendix C for details}.
We pre-trained the proposed EEG encoder with $N_{l_s}=6$, $N_{l_t}=6$, $N_{h_s}=8$ and $N_{h_t}=8$ \textit{using both the temporal and spatial modules}, with sequence length $t=64$ on contrastive learning objective, with hidden dimension $d=1024$ to match the latent space of stable diffusion.
\begin{table}
\small
\centering
\begin{threeparttable}
\caption{KNN and K-means clustering accuracies on the EEG-CVPR40 dataset for zero shot class generalization.}
\label{tab:unseen_cvpr40}
\begin{tabular}{lccc}
\toprule
\textbf{Model} & \textbf{SVM} & \textbf{KNN} & \textbf{K-Means}\\
\midrule
Cogni-Net \cite{Cogni-Net-8803717} & 0.78 & 0.72 & - \\
EEGStyleGAN-ADA \cite{singh2024learning} & 0.93 & 0.86 & 0.625 \\
\textbf{Ours} & -- & \textbf{0.997} & \textbf{0.699} \\
\bottomrule
\end{tabular}
\end{threeparttable}
\end{table}

\begin{table}
\small  
\centering
\caption{FID and IS scores on the EEG-CVPR40 dataset.}
\label{tab:fid_is_cvpr40}
\begin{tabular}{lcc}
\toprule
\textbf{Model} & \textbf{IS} & \textbf{FID} \\
\midrule
Brain2Image (GAN) \cite{kavasidis2017brain2image} & 5.07 & - \\
Neurovision \cite{khare2022neurovision} & 5.15 & - \\
DCLS-GAN \cite{fares2020brain} & 6.64 & - \\
EEGStyleGAN-ADA \cite{singh2024learning} & 10.82 & 174.13 \\
DM-RE2I \cite{zeng2023dm} & 12.55 & - \\
NeuroDM \cite{qian2024neurodm} & 15.89 & - \\
GWIT \cite{lopez2025guess} & 33.87 & 78.11 \\
\textbf{Ours} & \textbf{25.15} & \textbf{81.07} \\
\bottomrule
\end{tabular}
\end{table}


\section{Acknowledgment}
The authors declare no conflict of interest.
\bibliographystyle{elsarticle-num} 
\bibliography{ref}




\clearpage
\appendix

\section{Contributions of Temporal and Spatial Modules}
\label{sec:contribution-spatio-temporal}
To reveal the  significance of spatial and temporal modules of our proposed EEG Encoder~\ref{sec:Saption-Temporal-Encoder}, we experiment in the settings where only one of the two modules is used for feature extraction. We keep the sequence length fixed to $64$ for EEG-CVPR40 dataset~\ref{ds:eeg-cvpr40} and $32$ for ThoughtViz~\ref{ds:thoughtviz}. Further for each module, we also vary the number of encoder layers and self attention heads, to quantify the relationship between complexity of the architecture and the quality of the features extracted. The results are given in Table~\ref{tab:cvpr-abal-enc} and Table~\ref{tab:tviz-abal-enc} for EEG-CVPR40 and ThoughtViz dataset respectively.

For EEG-CVPR40 dataset, it can be seen that the validation k-means accuracy is not effected at all by the choice of the module in the proposed encoder. However in ThoughtViz dataset, the results are significantly better for same number of layers and heads if we use temporal module, as compared to the spatial one.
Which hints that the class level discriminative features are fairly pronounced along both spatial and temporal views of the signals in EEG-CVPR40, but only along temporal dimension for the case of ThoughtViz.

Moreover the effect of increase or decreasing the individual number of layers and heads is observed to be negligible, signaling that feature space approximated by few layered transformers, is structured sufficiently good, at least for the immediate task of clustering and subsequently for that of classification.
Based on these observations we opt to use only the temporal module, because it's performance is satisfactory across both datasets, in main experiments and limit the number of layer and heads with $N_{l_s}=2$ and $N_{h_s}=4$.

\begin{table} 
\centering
\begin{tabular}{l c c c c c}
\hline
Module & $N_{l_t}$ & $N_{h_t}$ & $N_{l_s}$ & $N_{h_s}$ &  K-Means Acc(\%) \\
\hline

Temporal & 2 & 2 & 0 & 0 & 0.992 \\
Temporal & 2 & 4 & 0 & 0 & 0.993 \\
Temporal & 4 & 2 & 0 & 0 & 0.994 \\
Temporal & 4 & 4 & 0 & 0 & 0.994 \\
Spatial  & 0 & 0 & 2 & 2 & 0.996 \\
Spatial  & 0 & 0 & 2 & 4 & 0.996 \\
Spatial  & 0 & 0 & 4 & 2 & 0.998 \\
Spatial  & 0 & 0 & 4 & 4 & 0.998 \\
Both     &    2 & 2 & 2 & 2 & 0.996 \\
Both     &    2 & 4 & 2 & 4 & 0.996 \\
Both     &    4 & 2 & 4 & 2 & 0.996 \\
Both     &    4 & 4 & 4 & 4 & 0.997 \\
\hline
\end{tabular}
\caption{Analyzing the change in K-means accuracy by adding or removing spatial and temporal modules in proposed encoder~\ref{sec:Saption-Temporal-Encoder}. The results are given for validation split of EEG-CVPR40 dataset.~\ref{ds:eeg-cvpr40}}
\label{tab:cvpr-abal-enc}
\end{table}

\begin{table} 
\centering
\begin{tabular}{ l c c c c c}
\hline
Module & $N_{l_t}$ & $N_{h_t}$ & $N_{l_s}$ & $N_{h_s}$ &  K-Means Acc(\%) \\
\hline
 Temporal & 1 & 1 & 0 & 0 & 0.764 \\
 Temporal & 1 & 2 & 0 & 0 & 0.758 \\
 Temporal & 2 & 1 & 0 & 0 & 0.765 \\
 Temporal & 2 & 2 & 0 & 0 & 0.767 \\
 Temporal & 2 & 7 & 0 & 0 & 0.777 \\
 
 Spatial  & 0 & 0 & 1 & 1 & 0.405 \\
 Spatial  & 0 & 0 & 1 & 2 & 0.366 \\
 Spatial  & 0 & 0 & 2 & 1 & 0.658 \\
 Spatial  & 0 & 0 & 2 & 2 & 0.661 \\
 Spatial  & 0 & 0 & 2 & 8 & 0.670 \\

 Both     & 1 & 1 & 1 & 1 & 0.604 \\
 Both     & 1 & 2 & 1 & 2 & 0.682 \\
 Both     & 2 & 1 & 2 & 1 & 0.749 \\
 Both     & 2 & 2 & 2 & 2 & 0.715 \\
 Both     & 2 & 7 & 2 & 8 & 0.745 \\

\hline
\end{tabular}
\caption{Analyzing the change in K-means accuracy by adding or removing spatial and temporal modules in proposed encoder~\ref{sec:Saption-Temporal-Encoder}. The results are given for validation split of ThoughtViz dataset.~\ref{ds:eeg-cvpr40}}
\label{tab:tviz-abal-enc}
\end{table}

\section{Effect of Signal Length on Feature Extraction}
\label{sec:time-analysis}

\begin{figure}[htbp]
    \centering
    \includegraphics[width=0.7\linewidth]{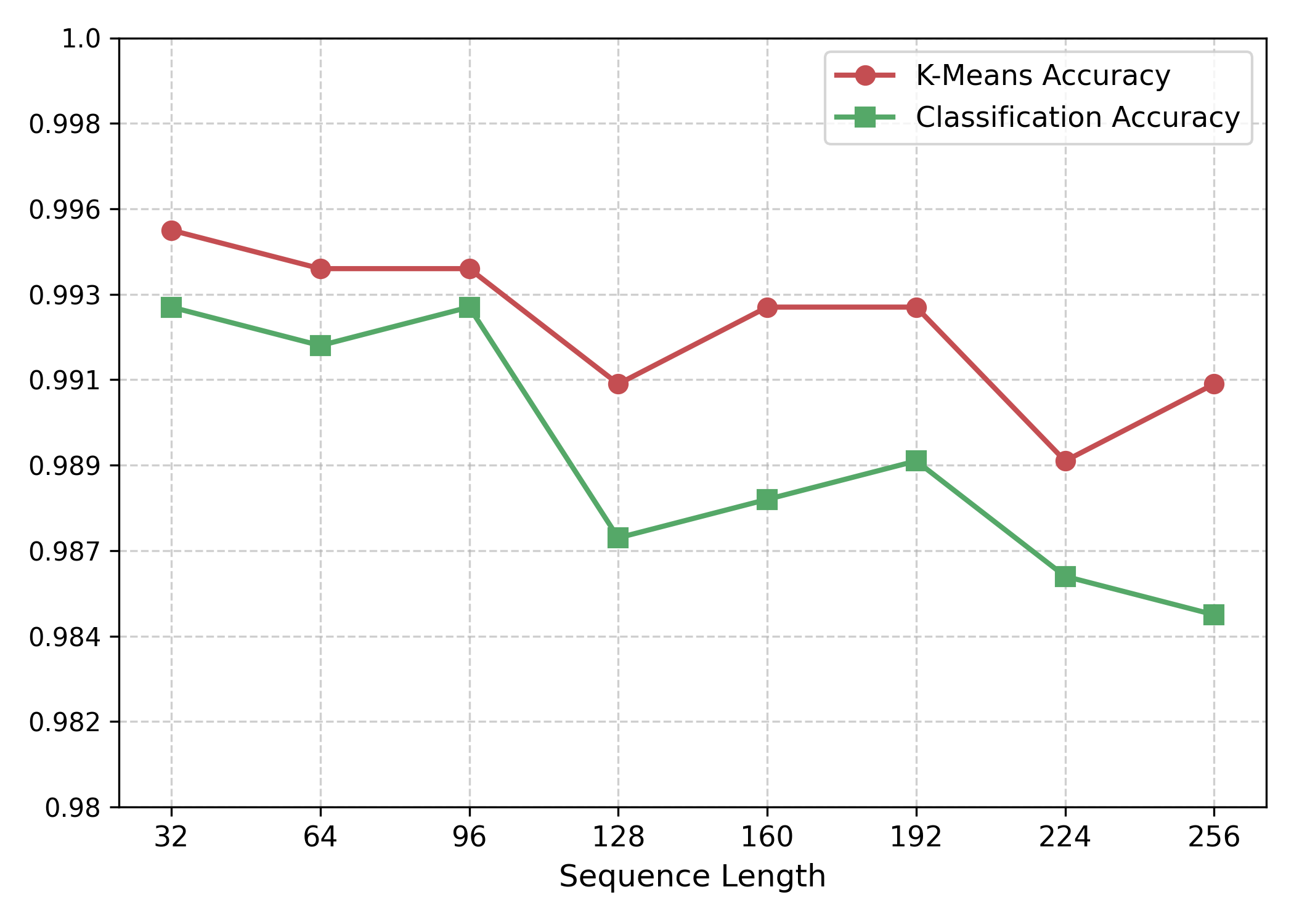}
    \caption{Effect of sequence length on performance of proposed encoder on EEG-CVPR40 dataset.}
    \label{fig:seq_len_accuracy}
\end{figure}

The duration of EEG signal required for BCI tasks, directly determines the required computation resources and consequently its application in real world scenarios. Therefore we have explored the effect of signal duration on model's performance to find the optimal duration for downstream tasks on EEG-CVPR40 dataset. We experimented with increasing signals duration starting from 32 to 256 time steps with identical hyper-parameters on contrastive learning task. We evaluated the performance of proposed encoder using K-means clustering accuracy and classification accuracy over validation dataset and found it's performance to be consistent even with reduced duration of signal, which shows the robustness of proposed methods.
The results are presented in Table~\ref{tab:duration_vs_acc}. For better visualization the trend is also given in Figure~\ref{fig:seq_len_accuracy}.

\begin{table}[h!]
\centering
\caption{K-means clustering and classification across different sequence lengths on EEG-CVPR40 dataset}
\label{tab:duration_vs_acc}
\begin{tabular}{c c c c c}
\toprule
\textbf{Seq. Len} &  \textbf{Classification Acc (\%) } & \textbf{K Means Acc (\%) } \\
\midrule
32  & 0.993 & 0.995 \\
64  & 0.992 & 0.994 \\
96  & 0.993 & 0.994 \\
128 & 0.987 & 0.991 \\
160 & 0.988 & 0.993 \\
192 & 0.989 & 0.993 \\
224 & 0.986 & 0.989 \\
256 & 0.985 & 0.991 \\
\bottomrule
\end{tabular}
\end{table}

\section{Effect of Encoder's Complexity on Image Generation}
\label{sec:enc-complexity-generation-analysis}
The image generation results reported in~\ref{res:img-gen-cvpr40}, use scaled up version of proposed EEG encoder~\ref{sec:Saption-Temporal-Encoder}, using both spatial and temporal modules. This choice is motivated by the empirical observation that increasing the encoders complexity, results in image generation with better FID and IS, even though the clustering accuracy remains similar, as explained in~\ref{sec:contribution-spatio-temporal}.

\begin{figure}[ht]
    \centering
    \includegraphics[height=0.80\textheight, keepaspectratio]{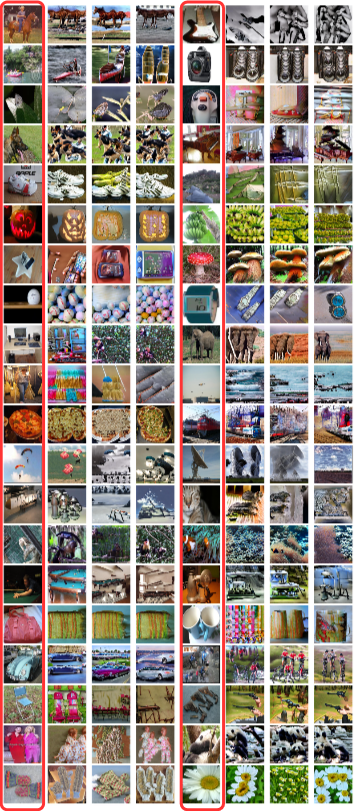}
    \caption{Images generated by Stable diffusion through conditioning from scaled up version of the EEG encoder, using both spatial and temporal modules. Columns 1 and 4  show the ground truth images, while the adjacent columns display the reconstructed images from the corresponding class.}
    \label{complex:generationResults}
\end{figure}

\begin{figure}[ht]
    \centering
    \includegraphics[height=0.80\textheight, keepaspectratio]{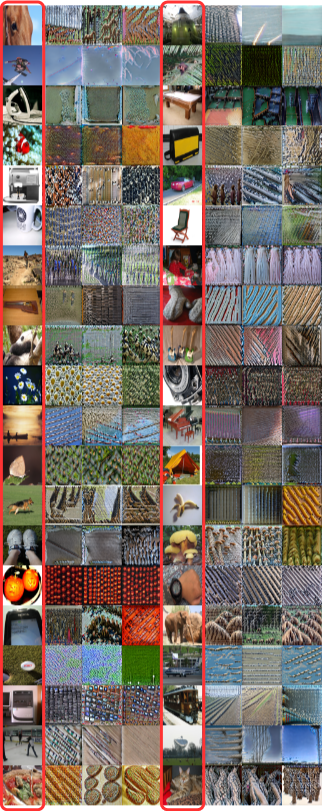}
    \caption{Images generated by Stable diffusion through conditioning from the temporal only module of the EEG encoder. Columns 1 and 4  show the ground truth images, while the the reconstructed images from the corresponding class.}
    \label{simple:generationResults}
\end{figure}

For illustration, we experiment by pretraining a low complexity encoder, with $N_{l_t}=2$, $N_{h_t}=4$ and $N_{l_s}=0$ \textit{using only the temporal module}, with sequence length $t=32$ on contrastive learning objective, with hidden dimension $d=1024$. We then use this pretrained encoder to finetune the Stable Diffusion in the same manner as explained in~\ref{res:img-gen-cvpr40}.
The resulting model's performance is far below than the previous one with, FID score of $245.93$ and IS of $9.47$, indicating the superiority of high representation abilities of the complex encoder with both spatial and temporal modules. Similar conclusion can be drawn through comparison of the visual quality of the generated images, when both temporal and spatial modules are employed, figure~\ref{complex:generationResults} and when only temporal module is employed, figure~\ref{simple:generationResults}.

\end{document}